# Natural Language Programming in Medicine: Administering Evidence Based Clinical Workflows with Autonomous Agents Powered by Generative Large Language Models

Akhil Vaid MD[1,2], Joshua Lampert* MD[1,2,3], Juhee Lee* MD[4], Ashwin Sawant* MD[1,5], Donald Apakama* MD[1,6,7], Ankit Sakhuja* MD[1,2,8], Ali Soroush MD[1,9], Sarah Bick MD[4], Ethan Abbott MD[1,6], Hernando Gómez MD[10], Michael Hadley MD[11], Denise Lee MD[1,12], Isotta Landi PhD[1,2], Son Q. Duong MD MS[1,13], Nicole Bussola PhD[1], Ismail Nabeel MD[14], Silke Muehlstedt[15], Robert Freeman RN[1,16], Patricia Kovatch BS[17], Brendan Carr MD MS[6], Fei Wang PhD[18,19], Benjamin S. Glicksberg PhD[2], Edgar Argulian MD[11,20], Stamatios Lerakis MD PhD[11,20], Rohan Khera, MD, MS[21,22,23], David L. Reich MD[24], Monica Kraft MD[5], Alexander Charney MD PhD[1,25], Girish Nadkarni MD MPH[1,2,26]

*Contributed equally

**Corresponding author**
Akhil Vaid
akhil.vaid@mssm.edu

**Affiliations**

1. The Charles Bronfman Institute for Personalized Medicine, Icahn School of Medicine at Mount Sinai, New York, New York, USA
2. The Division of Data Driven and Digital Medicine (D3M), Department of Medicine, Icahn School of Medicine at Mount Sinai, New York, New York, USA
3. Helmsley Electrophysiology Center, Mount Sinai Hospital, New York, New York, USA
4. Division of Genetics and Genomics, Boston Children's Hospital, Harvard Medical School, Boston, Massachusetts, USA
5. Department of Medicine, Icahn School of Medicine at Mount Sinai, New York, New York, USA
6. Department of Emergency Medicine, Icahn School of Medicine at Mount Sinai, New York, New York, USA
7. Institute for Health Equity Research, Icahn School of Medicine at Mount Sinai, New York, United States
8. Department of Critical Care Medicine, Icahn School of Medicine at Mount Sinai, New York, New York, USA
9. Department of Gastroenterology, Icahn School of Medicine at Mount Sinai, New York, New York, USA
10. Department of Critical Care Medicine, University of Pittsburgh School of Medicine, Pittsburgh, PA, USA
11. Mount Sinai Heart, Icahn School of Medicine at Mount Sinai, New York, New York, USA

12. Department of Surgery, Icahn School of Medicine at Mount Sinai, New York, New York, USA
13. Division of Pediatric Cardiology, Department of Pediatrics, Icahn School of Medicine at Mount Sinai, New York, New York, USA
14. Department of Environmental Medicine and Climate Science, Icahn School of Medicine at Mount Sinai, New York, New York, USA
15. Hasso Plattner Institute at Mount Sinai, Icahn School of Medicine at Mount Sinai, New York, New York, United States
16. Institute for Healthcare Delivery Science (IHDS), Icahn School of Medicine at Mount Sinai, New York, United States
17. Department of Scientific Computing, Icahn School of Medicine at Mount Sinai, New York, New York, USA
18. Department of Population Health Sciences, Weill Cornell Medical College, Cornell University, New York, New York, USA
19. Institute of Artificial Intelligence for Digital Health, Weill Cornell Medical College, Cornell University, New York, New York, USA
20. Department of Cardiology, Icahn School of Medicine at Mount Sinai, New York, New York, USA
21. Section of Cardiovascular Medicine, Department of Internal Medicine, Yale School of Medicine, New Haven, Connecticut, USA
22. Section of Health Informatics, Department of Biostatistics, Yale School of Public Health, New Haven, Connecticut, USA
23. Center for Outcomes Research and Evaluation, Yale-New Haven Hospital, New Haven, Connecticut, USA
24. Department of Anesthesiology, Perioperative and Pain Medicine, Icahn School of Medicine at Mount Sinai, New York, USA
25. Department of Psychiatry, Icahn School of Medicine at Mount Sinai, New York, New York, USA
26. Department of Nephrology, Icahn School of Medicine at Mount Sinai, New York, New York, USA

# Abstract

## Background

Generative Large Language Models (LLMs) have emerged as versatile tools in healthcare, demonstrating the ability to regurgitate clinical knowledge and pass medical licensing exams. Despite their promise, they have been largely treated as slow, imperfect information retrieval tools and face limitations such as data staleness, resource intensity, and manufacturing incorrect text – reducing their applicability to dynamic healthcare settings.

## Methods

This study explored the functionality of both proprietary and open-source LLMs to act as autonomous agents within a simulated tertiary care medical center. Real-world clinical cases across multiple specialties were structured into JSON files and presented to agents for solution using the resources available to a human physician. Agents were created using LLMs in combination with natural language prompts, tools with real-world interactions, and standard programming techniques. The technique of Retrieval Augmented Generation was used to provide agents with updated context whenever appropriate. Expert clinicians collected and evaluated model responses across several performance metrics including correctness of the final answer, judicious use of tools, guideline conformity, and resistance to hallucinations.

## Findings

Agents showed varied performance across specialties, with proprietary models (e.g., GPT-4) generally outperforming open-source models. The use of Retrieval Augmented Generation (RAG) improved guideline adherence and contextually relevant responses for the best performing model.

## Interpretation

LLMs can effectively function as autonomous agents in healthcare by leveraging their generative capabilities and integrating with real-world data. The study highlights the potential of LLMs to enhance decision-making in clinical settings through tailored prompts and retrieval tools. However, the variability in model performance and the necessity for ongoing manual evaluation suggest that further refinements in LLM technology and operational protocols are needed to optimize their utility in healthcare.

## Introduction

Generative Large Language Models[1] (LLMs) have garnered significant attention due to their advanced multi-modal, conversational capabilities. Being able to program new behavior into such models using natural language prompting gives them additional flexibility over traditional, task-specific machine learning methods. Such flexibility may be harnessed downstream for a variety of healthcare-associated tasks such as being conversational assistants[2], visual question-answering[3], and document classification[4]. However, LLM utilization has comes with challenges. By themselves, LLMs function as slow, imperfect databases that, when queried directly, may produce erroneous or fabricated responses—phenomena commonly referred to as "hallucinations."[5] Additionally, both the creation and operation of these models are extremely resource-intensive[6]. Another important concern is that the knowledge contained within a model may grow stale or may not cover specific situations. While partially mitigating this is possible by manual insertion of verified information at time of inference[7], or through supervised fine-tuning[8], either option is slow, time-consuming with difficulties scaling.

For healthcare use cases[9], LLMs have been shown to be able to regurgitate clinical knowledge[10], and pass medical licensing exams[11]. However, such tests do not capture the complex dynamics of real-world healthcare settings. General-purpose models cannot consider major differences in institutional preferences or knowledge, individual patient profiles, and demographic factors. Furthermore, the structured nature of medical question-answer tasks does not reflect the breadth of decision-making freedom that clinicians possess. Consequently, the quintessential clinical question of *"What is the next best thing to do for this patient?"* cannot be effectively addressed by traditional question banks used to assess trainee knowledge.

Despite these limitations, LLMs excel in context-based quasi-reasoning due to their capability for generating subsequent tokens following provided text. This capability can be leveraged to create autonomous agents[12] capable of operating within complex environments given an initial set of operating instructions. One such application is in the practice of evidence-based medicine (EBM)[13,14], where adherence to defined protocols can streamline clinical tasks. In EBM, clinical encounters typically follow a structured approach that includes interviews, physical examinations, differential diagnoses, and the sequential use of lab tests and imaging studies before therapeutic recommendations. The sequence of these actions, dictated by both clinician experience and patient-specific factors, must also accommodate [15] rapid changes in medical guidelines, standards of care, and an ever-expanding body of research.

The conceptualization of LLM based agents[16] as tiered entities—comprising the core model and an additional layer of natural language that imposes behavioral constraints—presents a transformative approach to managing the complexity of modern medical practice. By shifting from a paradigm of Natural Language Processing[17] to one of Natural Language Programming, agents can provide refined, context-aware assistance to clinicians. This paradigm not only enhances the utility of LLMs but also signals a broader change in how we implement artificial intelligence in critical, information-dense fields like medicine and beyond.

Recent work has explored the abilities of LLMs as agents that can extract patient history in a conversational manner[18], as well as attend to tasks of fetching information from databases[19] by acting as a translational layer between natural language and database-specific query languages. Prior attempts[20] at demonstrating the usefulness of agents as autonomous solvers of questions pertaining to the practice of medicine have been limited by and to non-expert and non-granular (automated) evaluation of simpler models, and limitations in prompting and response interpretation.

In this work, we explore the ability of both proprietary and open-source LLMs to function autonomously within an agent-based workflow in an environment representative of a tertiary care medical center **(Figure 1).** We equip models with all the tools and effectors that are available to a physician, and measure how currently available models function when provided with a general-purpose set of instructions to adapt to evolving clinical scenarios. We demonstrate the use of techniques such as Retrieval Augmented Generation[21] (RAG) that can be used to automatically supplement inherent LLM knowledge and agent operation. We establish and evaluate the correctness and applicability of agentic responses through manual review by expert clinicians. Finally, we discuss how Natural Language Programming has, in essence, replaced Natural Language Processing as the dominant paradigm when dealing with complex systems that operate through transmission of the written word.

## Methods

### Data sources

We utilized real-world clinical cases seen by expert faculty clinicians across multiple specialties, namely cardiology, critical care, emergency medicine, genetics, and internal medicine (**Supplementary Table 1**). We selected representative cases for scenarios a physician would encounter within their practice. Each of these cases was then structured as a JSON[22] (JavaScript Object Notation) file containing discrete categories of clinically relevant information (keys) paired to associated values **(Supplementary Figure 1)**. We chose the JSON format for its ability to express the hierarchical organization of complex relationships within clinical cases, ease of human readability, and its structured nature that facilitates straightforward machine retrieval of values associated with known keys.

Each case also contained links to prevailing diagnostic and treatment guidelines. Guidelines were instituted as plain text and divided into two subheadings: initial assessment, and initial treatment. Where relevant, available documentation was also supplemented with imitation institutional guidelines for certain cases to simulate how a clinician must alter their recommendations in special circumstances.

Finally, each file contained a question for the LLM to answer given clinical context. For the purposes of this study, this question was always "*What is the next best step in management?*" since an appropriate answer would attend to most of clinical protocol – knowing which investigations to order, how to interpret results, generation of a diagnosis, and next best steps on top.

### Agents

An agent may be envisioned as a combination of four components. One, an underlying LLM. Since all generative LLMs operate in much the same manner, the actual model is interchangeable. Two, the text presented to the model, or prompt[23]. Prompts may be further divided into *system prompts* which are prepended to all interactions with a model for the purpose of directing the LLM's manner of behavior; and a *user prompt* that contains context specific instructions and/or data for a model to analyze. Three, tools – combinations of text descriptors inserted into the system prompt, coupled to real-world effector functions that allow the model to interact with available infrastructure. Four, traditional programming techniques such as loops and regular expression-based searches to start, continue, and end the operation of an agent when certain set criteria are achieved. **(Figure 2)**

### Retrieval Augmented Generation (RAG)

RAG is a technique that enhances the capabilities of LLMs by combining their generative capabilities with the information retrieval abilities typical of search engines[21]. RAG can fetch relevant information from a vast dataset or external knowledge base before generating a response. This method helps in producing more accurate, informed, and contextually relevant outputs, especially when the internal knowledge of a language model might be outdated or insufficient. RAG use typically involves setting up an external source of curated information[24] that can be queried using either similarity-based metrics, adapters that translate between conventional database queries and natural language, or directly with natural language. Our implementation of RAG involved tool-based retrieval of case-specific guidelines stored within text files which were only shown to the agent if it arrived at the most likely correct diagnosis according to the curating clinician. In a separate set of evaluations, we disabled the RAG tool and evaluated how the best performing model might function when generating next steps without extraneous guidelines available.

**Tool creation**

We created tools allowing agents to retrieve information from cases using pre-defined keys. Models were required to explicitly ask for the results from a tool similar to a clinician ordering an investigation within a hospital setting. These tools were dedicated to retrieval of either of the patient's symptoms, signs (physical examination), past medical history, electrocardiogram (ECG), results from other machine learning models, laboratory studies, and imaging studies. **(Figure 1, Table 1)**. Tools for symptoms and signs worked without any input data, and returned long-form patient history, symptomatology, and physical exam findings respectively. In contrast, both the lab investigation and imaging study tools required the agent to provide a contextual input selected from a list of investigations. For example, "SERUM BILIRUBIN", or "CHEST X-RAY". The full list of available investigations was created by pooling the names of tests across all available cases to more closely replicate a real hospital setting. **(Supplementary Table 2)**

If an agent asked for a laboratory or imaging study not mentioned in the case, it was instead provided with a "Not available" text. Similarly, if there were no specific guidelines associated with the case or the agent provided an incorrect diagnosis to the RAG tool, it was provided with text stating "No updated guidelines available. Use your best clinical judgment". Finally, if the agent made an error in selecting a tool, it was provided with feedback stating "*<selection> is not a valid tool. Please try with one of <tool names>*". At this point, the agent was expected to modulate its output according to provided instructions in order to better use available tooling.

**Agentic operation**

A key differentiator between LLMs and other kinds of machine learning models is that LLM behavior is not deterministic even in highly structured scenarios[5,25]. Thus, prompt engineering[26] refers to the (often subjective) process of formulating and modifying prompts such that the downstream LLM generates text according to specification.

We formulated our system prompt as a set of instructions giving the agent an identity ("…as a professor of medicine"), and a general set of instructions about how to utilize provided tools. **(Figure 3, Full prompt detailed in Supplementary Table 3).**

These instructions recommended judicious use of available tools in a logical manner. Additionally, the system prompt contained instructions to treat the output of a tool as the starting point for an observation that would act as an internal monologue and recommend an action that would be the next tool in line that it would utilize. The agent was directed to stop using tools and provide a final answer once it thought itself confident of it, or when it thought that the tooling would no longer provide any useful information. As above, the overall output of the model was parsed at each step using a rule-based approach to see if the model had reached the final

answer – and if so, the execution of the process was halted, and the overall chain of responses logged for evaluation. This method of iteratively building the overall input to the model is described as chain-of-thought[27,28] prompting, and it allowed us to establish a context for the model to operate off and keep generated text on guardrails. Overall prompt development was performed in an iterative manner using ChatGPT-3.5.

## Performance evaluation
Another key difference between traditional classification or regression-based machine learning and generative modeling is the quantifiability of the outcomes. Well established metrics such as Areas Under the Curve, or Mean Square Error allow for precise, automated measurement of how well a traditional model might perform. However, evaluation of generative approaches is constrained to be subjective – especially within the context of a complex task such as patient management. Additionally, generated text must be manually evaluated by clinicians familiar with both the patient, as well as the conditions prescribed by the hospital system.

We tested multiple models within the agent framework as an exploration of how the currently available crop of LLMs may perform if allowed autonomous operation with the same set of operating instructions. Tested models were of both open-source (LLaMA 2-70B, Mixtral 8x7B, Meditron 70B) and proprietary (GPT-4, Gemini Pro, and Claude 3 Opus) provenance. Except for one open-source model (Meditron 70B) that was trained on medical text, all models were trained on general datasets. Each agent was allowed one attempt to generate coherent text related to a case. Runs were automatically evaluated to zero in case of deviations from the instructions put down as part of the system prompt - including hallucinating the history, not querying any tool, and/or jumping directly to a hallucinated final answer. All text was logged and manually evaluated by two clinicians of the same specialty in a single-blinded manner, i.e. the clinician was unaware of which model's responses they were looking at.

Responses were evaluated on a ten-point scale for correctness of final answer; judicious use of tools; conformity to established clinical guidelines in the final recommendation; and resistance to hallucinations. In case an agent was incorrect in its diagnosis, it was automatically marked 0 for the remaining tooling and guidelines as well.

Case difficulty and the amount of information needed to make useful decisions may vary contingent upon clinician experience and expertise. Therefore, we included a secondary appraisal of case difficulty as it would apply to a human physician into the evaluation process. Cases were graded for difficulty on a scale of ten as before. All cases are available for review within the linked online repository.

## Software and Hardware
We utilized the transformers[29], PyTorch[30], and LangChain libraries to create this framework and resulting agents. These libraries were called from within the Python[31] programming language (3.12.x).

Proprietary models are accessible over the internet through APIs and for these models, response generation was done securely in a cloud computing framework. The open-source models were run at 16-bit precision within the supercomputing cluster at the Icahn School of Medicine at Mount Sinai.

# Results

## Overall Performance

We evaluated agents based on three proprietary and three open-source LLMs across five medical specialties: Cardiology, Critical Care, Emergency Medicine, Genetics, and Internal Medicine. The evaluation was conducted by two clinicians from each specialty. We focused on three key metrics to assess their performance: correctness of the final answer, judicious use of tools, and conformity to established guidelines. Additional scoring was done for resistance to hallucinations in generated responses, according to case difficulty as reported by evaluating clinicians, and for conformity to guidelines with RAG disabled for the best performing agent. Scores reported below are averages of all five cases within a specialty and fall within a scale of 1 to 10.

## Comparative Analysis Across Metrics by Specialty

Correctness of the Final Answer: The agents based on GPT-4 and Claude 3 Opus often scored similarly and outperformed the agent based on Gemini Pro in correctness across most specialties. Both the GPT-4 and Claude agents achieved a perfect score of 10 in Critical Care, while the highest score the Gemini Pro-based agent reached in any category was 5·3, also in Critical Care. This was largely due to the Gemini Pro agent's inability to follow instructions as provided within the system prompt, leading to abrupt terminations or direct jumps to an incorrect final answer. **(Figure 4, Table 2)**

Judicious Use of Tools: The GPT-4-based agent consistently scored high in judicious use of tools, surpassing other agents in Internal Medicine with a score of 8·6 and matching the Claude 3 Opus-based agent in Critical Care with a score of 8·1. The Claude agent showed strong performance in Cardiology and Internal Medicine with scores above 7. The Gemini Pro-based agent's performance in tool usage was considerably lower, not exceeding 3·9 in any specialty. **(Figure 4, Table 2)**

Conformity to Guidelines: The GPT-4-based agent typically led with respect to conformity to guidelines, scoring highest in Internal Medicine (8·2) and Critical Care (8·9). The Claude-based agent also displayed strong adherence, particularly in Cardiology (8·1) and Genetics (8·4). The Gemini Pro-based agent's highest conformity score was 3·5 in Cardiology, indicating a significant gap compared to the other agents. **(Figure 4, Table 2)**

Resistance to Hallucinations: Across completed cases, all agents demonstrated excellent resistance to hallucinations. Scores were perfect for agents built using all three proprietary models according to both evaluating clinicians. **(Supplementary Figure 2)**

## Retrieval Augmented Generation

The agent based on GPT-4 achieved an average performance uplift of 20% across all specialties in terms of conformity to guidelines when using RAG. The most substantial change (52%) was seen for Internal Medicine cases (5·4 to 7·2). These results suggest that additional direction supplied by the presence of question-specific context is sufficient to make generation more use-case appropriate. **(Figure 5, Table 3)**

**Performance by Difficulty**

Of cases judged to be "hard" difficulty by evaluating clinicians, the GPT-4-based agent had highest performance across all metrics – exceeding performance even at cases established to be of "easy", or "medium" difficulty. The GPT-4 agent received perfect scores (10·0) for hard cases for correctness, around 9 for judicious use of tools, and guideline conformity. The Claude and Gemini Pro-based agents came in second and third, respectively. Performance for agents based on both these models was seen to be inversely proportional to the difficulty of the case, somewhat mimicking the likely performance of a human practitioner. The Gemini Pro-based agent had difficulty negotiating the medium and harder difficulty cases, with scores not exceeding 4·0 for any metric. **(Supplementary Figure 3)**

**Open-source models**

Agents based on tested open-source models performed considerably worse than those based on proprietary models. The LLaMA-2 70B-based agent managed to complete cases only in Critical Care, scoring a 3·8 in correctness of the final answer, but it failed to address any cases successfully in other specialties including Cardiology, Emergency Medicine, Genetics, and Internal Medicine, scoring zero across all metrics· The Mixtral 8x7B-based agent also showed poor performance, unable to complete any cases in Cardiology and Internal Medicine, and only managing marginal scores in Critical Care (2 in correctness), Emergency Medicine (2·9 in correctness), and Genetics (3·4 in correctness). The agent built on Meditron 70B, a medical fine-tune based on LLaMA-2 70B, did not complete a single case across all specialties, scoring zero in all metrics. **(Supplementary Table 4, Supplementary Figures 4 and 5).**

## Discussion

In this study, we demonstrate that generative Large Language Models (LLMs), real-world effector functions, and parsing of natural language responses can be combined for the creation of autonomous agents. We show that knowledge embedded within these models can be utilized downstream to make LLMs emulate physician behavior in solving clinical cases across disciplines. Additionally, we demonstrated the utility of RAG to make agents more suited towards operation in settings requiring a high degree of institutional knowledge or that demand application of updating information. We develop and open-source general natural language instructions for this functionality and tested the resulting framework against currently available proprietary and open-source LLMs. Given the current state of clinical decision support systems, agentic operation may represent a more consolidated approach in terms of how to approach the patient, as well as a better way to get to the all-important next-best-step of management, especially in resource or experience limited settings.

We found that across chosen metrics, the agent based on GPT-4 consistently demonstrated high performance, indicating its superior capacity for integrating accuracy, appropriate tool use, and strict guideline adherence. The agent created using Claude 3 Opus also performed well but was occasionally outmatched by the GPT-4 agent, especially in tool usage and guideline adherence. The Gemini Pro-based agent's lower scores were mostly due to an inability to complete provided clinical scenarios. Performance trends persisted across case difficulty. Agents based on open-source models performed considerably worse and were only able to complete a minimal number of cases.

In the absence of standardized benchmarks, it is important that the capabilities of such agents be manually evaluated on real-world data and scenarios. Automated evaluation against synthetic data cannot and does not consider the importance of institution specific effects, step-by-step reasoning that leads the model to an answer, or context-aware and reasonable resource utilization. While the more complex models tested were able to achieve consistently good performance, systemic biases in automated evaluation protocols, and overtly rigid prompting can lead to different conclusions[20].

While all Large Language Models (LLMs) are exposed to vast amounts of text, including medical textbooks—which theoretically should equip them to handle commonly seen pathologies effectively—we observe substantial variations in their capabilities. Traditionally, evaluation metrics for these models have emphasized the semantic and grammatical accuracy[32] of generated text, as well as their proficiency in solving multiple-choice questions. However, these metrics do not capture the practical effectiveness of LLMs in real-world applications. We argue that the steerability of a model, although more subjective, is a more useful indicator of its utility in dynamic and complex environments. This is evidenced by the more complex models tested within this framework being able to achieve not just excellent diagnostic performance, but also being able to generate the basis of future personalized care for each patient. In comparison, several models that have equaled or exceeded GPT-4 performance on individual benchmarks[33,34] (including clinical information retrieval) were found to perform much worse, or not be able to respond to complex instructions at all.

A cornerstone of this agentic operation is model agnosticism. Our implementation allows for the underlying model to be switched out by changing one line of code, thereby allowing near-instantaneous performance improvements as better, more steerable models become available. This applies to both proprietary and open-source solutions. In this context, we stipulate that existing open-source modeling efforts may not be considered suitable for autonomous operation, but the utilization of proprietary models will always come with concerns of patient privacy, cost and difficulty tracking decision making since the underlying LLM may be subject to change or update without notice. Conversely, for open-source solutions the problems are initial investment into infrastructure, limited amount of context that a model can successfully process, and demonstrably worse, unstable performance.

Properly vetted agents can serve pivotal roles across various scenarios: as triage specialists in emergency departments, managing intake and initial assessments, or as first points of contact in outpatient settings, collecting preliminary patient histories. In all these cases, this could lead to workload reduction, and optimized care. Fully autonomous models also represent other advantages. As above, minor prompt engineering enables models to output all differential diagnoses at each step of the process, and the reasoning behind them. Such directions may be immensely useful for clinical education since they represent step-by-step considerations of the next-best-step specific to the patient– something that may be difficult to glean from mainstream textbooks. Additionally, clinicians have access to the entire thought-process of the model through the "observations" contained within the output. Thus, the manner of operation demonstrated is transparent and in keeping with clinical guidelines. In contrast, clinical decision support systems must still provide complex saliency maps, or measures of importance that do not support any claims of causality[35]. Further, LLMs operating in this manner may also largely circumvent the concerns that issue with the use of predictive models in healthcare settings[36].

LLM operation must also be considered in terms of feasibility. The costs of training such models ensures that new knowledge cannot be easily injected into such models. This is an especially important concern for applications which rely on domain specific knowledge or where such knowledge changes often. A relevant example is antibiotic resistance[37]. Microbial populations and susceptibilities change over time and geography. It stands to reason that such knowledge cannot be made part of a model, or it will quickly start to issue incorrect recommendations. However, institutions already maintain records of antibiotic resistance which LLMs can parse to issue very specific recommendations. As we demonstrate, even for the best available model (GPT-4), the addition of RAG increased performance at generating text that conformed to the best available clinical knowledge **(Figure 5, Table 3)**. Prompt engineering[26] shortens the time required to program new behaviors with changing circumstances, or when errors are identified. For example, an agent may order tests which are either too expensive or considered unnecessary. Prompt engineering would allow for remedying situations like these simply by telling the model to "only perform absolutely necessary investigations", or by making the model aware of how long results may take. Similar considerations may be applied to clinical guidelines as well. We believe the best way to engineer these prompts will vary from facility to facility and should be an avenue for future investigation.

Finally, we also show iterative construction of the model prompt reduces hallucinations. By utilizing chain-of-thought prompting coupled to stepwise injection of verified information, we steer towards a specific direction and discard incorrect assumptions. This is an especially pertinent approach over utilizing models as pure question-answering tools since hallucinations may snowball[38]. The effectiveness of this approach is demonstrated by the lack of hallucinations in the responses of even otherwise poorly performing models. Additionally, models which operate purely within chat-interfaces are subject to the possibility of "jailbreaking"[39] – wherein malicious natural language instructions can override the model's operating instructions. Agentic operation reduces or entirely removes human input for operation, and largely reduces such misuse potential-especially useful for information-controlled hospital settings.

Our work must be considered alongside its limitations. First, curation and translation of cases from the electronic healthcare record to JSON files needed human intervention, as well as subsequent evaluation of long-form LLM responses. Therefore, we were limited in terms of the overall sample size of the study. However, cases were carefully chosen to be representative of what a clinician operating within a specialty might expect to encounter. As before, this method of evaluation is also important to consider many aspects of agent-based operation that cannot be attended to with automated testing. Second, while case data was deposited within machine readable JSON files, any extension of this work to an actual clinical setting would involve setting up adapters that bridge the infrastructure of the clinical setting with the tools available to the agent. Finally, clinical guidelines for certain conditions were especially voluminous, and we were restricted to GPT-4 generated summaries of these guidelines to fit them into reasonably sized prompts. We attempted to circumvent this limitation by asking models to generate their final recommendations using a combination of their own clinical knowledge and provided text; but we stipulate the larger models may be able to generate bespoke recommendations to individual patients either with more context, or more sophisticated guideline retrieval. Finally, remaining limitations issue from the underlying models. As before, the choice between open-source and proprietary models brings caveats on either side, and which model to employ as the driver is subject to institutional and situation specific concerns. Metrics as shown conform to the LLaMA 2 class of models and derivatives, while the LLaMA 3 class of models has recently been released. These models have generally displayed better performance across a variety of synthetic benchmarks and may be utilized within this framework through minimal modifications of the source code.

In conclusion, LLMs are profound tools that have wide-ranging implications in how we consider the deployment of AI systems in healthcare. Operationalizing them as agents allows them to integrate with the moving parts of a healthcare system – a paradigm change in approach that could lead to significant changes in how evidence-based medicine performed by providers and experienced by patients.

**Figures**

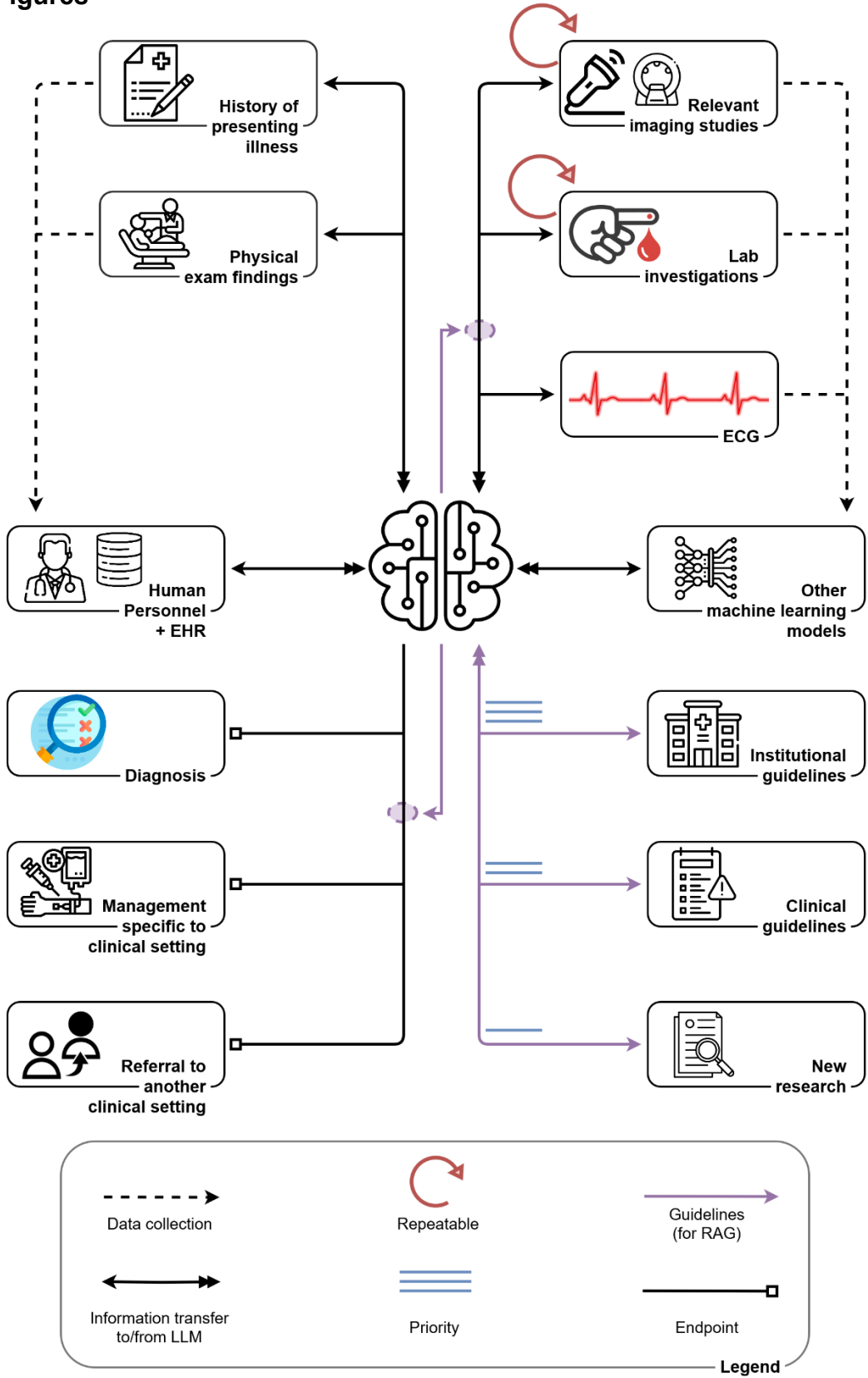

**Figure 1. Workflow**

The executor LLM relies on human personnel to generate the history of presenting illness, and the physical exam findings. Following this, it autonomously chooses which lab investigations or imaging studies to order, and how to interpret them. These interpretations form the basis of the model's next step or order. During this process, it may also query the output of existing machine learning systems.

Following this, the model queries all available literature on the most likely differential diagnosis it is considering. In this figure, blue horizontal lines indicate the degree of importance given by the model to any one source of information. This is an important consideration since we want institutional knowledge to override general purpose guidelines which may not be entirely appropriate for any one patient population.

Finally, the model issues its recommendations based entirely on the question it is tasked to answer. This may be generating the final diagnosis, the management specific to the clinical setting it finds itself in, or referral to another facility.

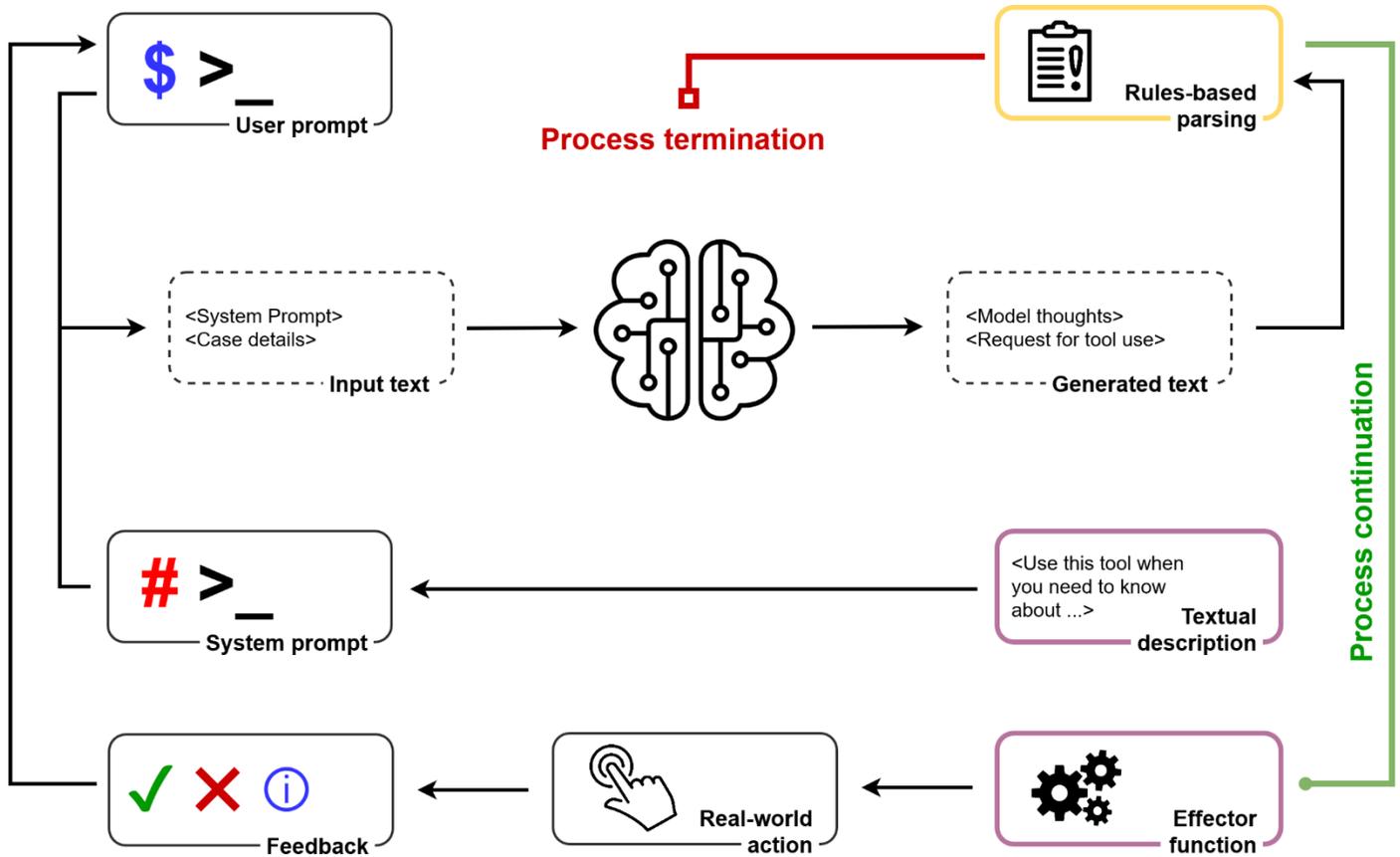

**Figure 2. Agentic operation**

The agentic workflow depicted in the figure outlines a structured interaction process between user prompts and an LLM agent-based system. The process initiates with a user prompt (represented by the blue terminal icon), which is parsed and formulated into input text comprising system prompts and case details. This input is then processed by the LLM (symbolized by the central brain icon) that generates text containing model thoughts and requests for tool use. Tools are shown in purple boxes as two components – textual descriptions and attached effector functions.

Upon generation of text, two possibilities emerge. The first is **Process Termination** (indicated by the red line): If the generated text signifies the end of the process upon rules-based parsing, the workflow concludes. The second is **Process Continuation** (illustrated by the green line): If the generated text calls for a tool, it triggers the use of an effector function. This function executes real-world actions (depicted by the hand icon), feedback from which are provided as text, and fed back as the user prompt, thereby creating a loop that continues until process termination.

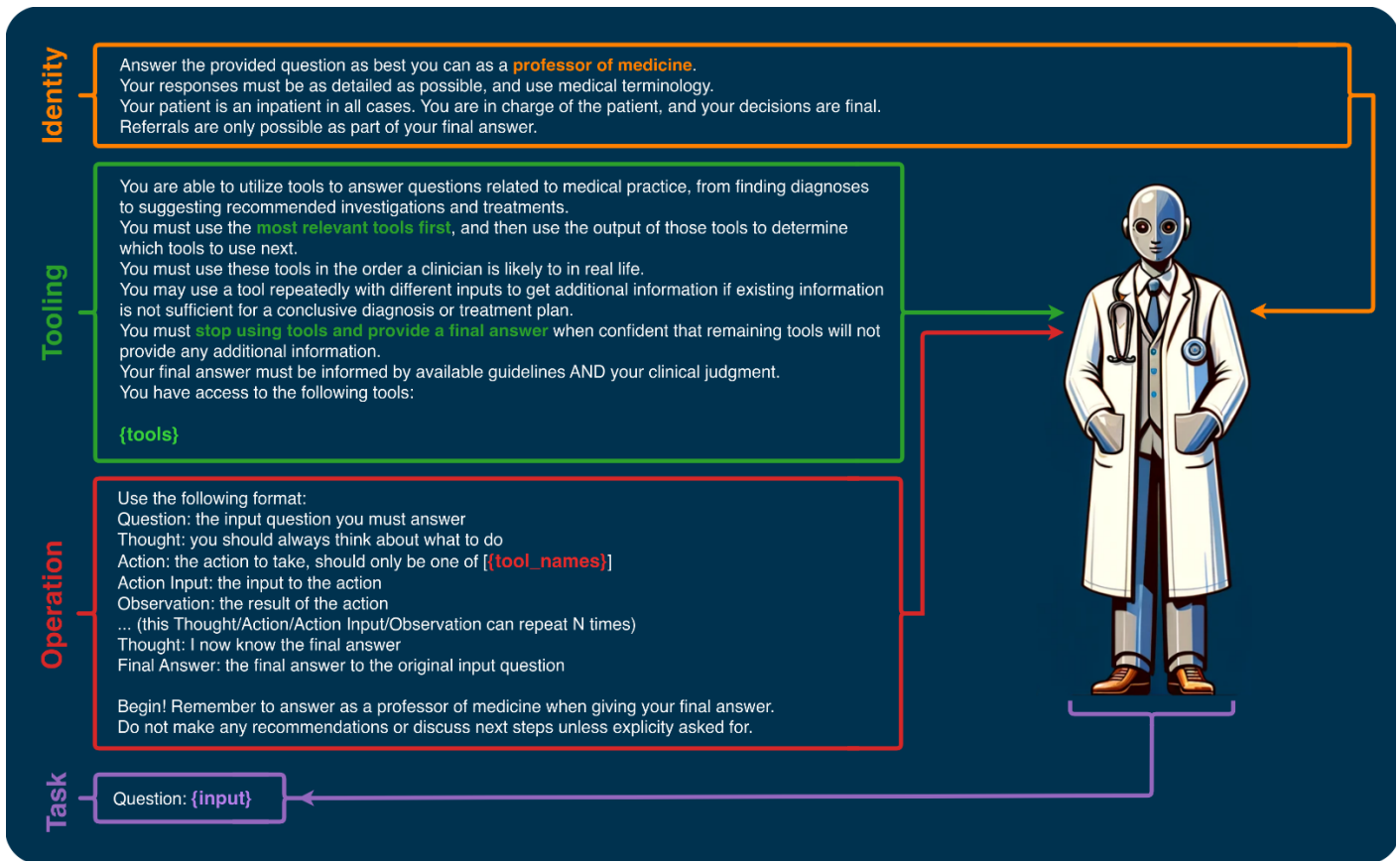

**Figure 3. Initial (system) prompt**

Preamble text that establishes the framework for the LLM. Overall text may be separated into 4 parts each with their own function:

(1) Identity: Tells the model who it is and how it's expected to respond. The model was asked to assume the identity of a "professor of medicine" for most of our testing.

(2) Tooling: Provides the model with a general sense of directions regarding when to use tools. Actual choice and order of tool use is up to the model. Also establishes a stop condition for the model i.e. when it reaches a final answer.

(3) Operation: Establishes "chain-of-thought" prompting structure. The narrative the model uses to generate the next word is built up piecemeal through repeated iterations of thoughts, actions, and observations in that order until the final answer is reached.

(4) Task: What the model is supposed to provide a solution for - "*What is the next best step in management*"?

Text in curly brackets is replaced by names of tools and the question at runtime. The full prompt so generated is available in **Supplementary Table 3**.

# Performance by specialty

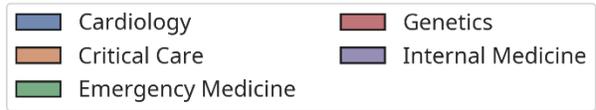

## a Correctness of final answer

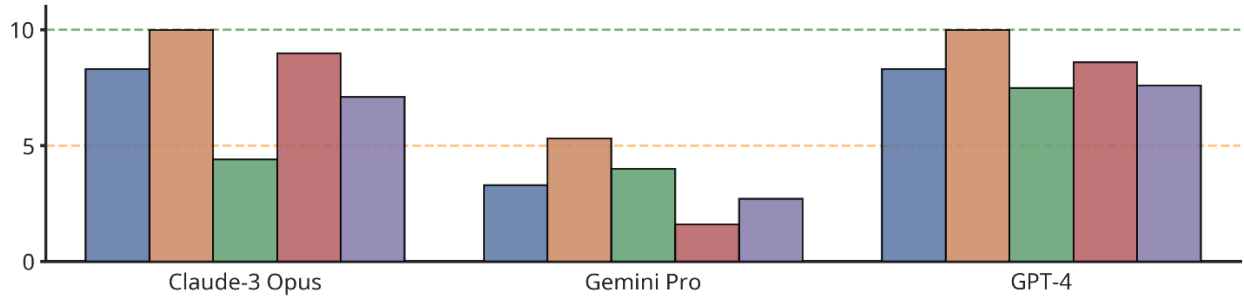

## b Judicious use of tools

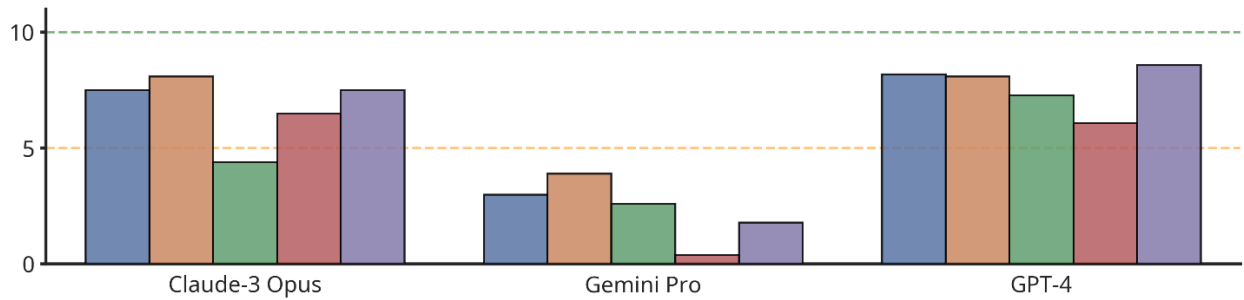

## c Conformity to guidelines

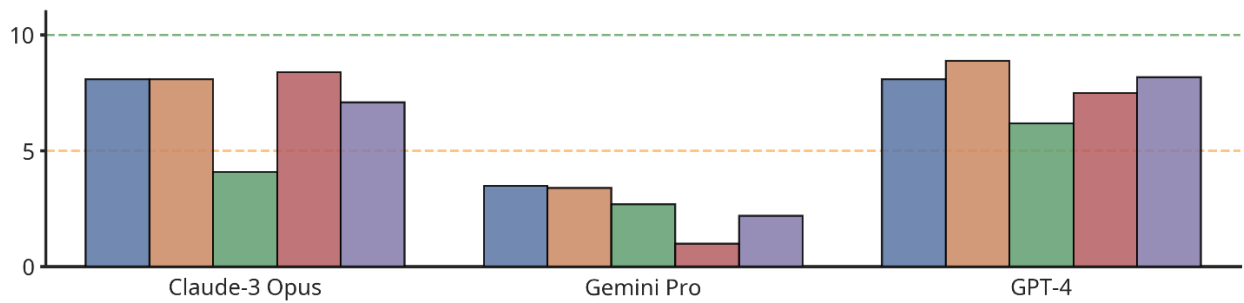

**Figure 4**. **Proprietary model performance within framework by specialty**
All models (Gemini Pro, Claude 3 Opus, and GPT-4) had Retrieval Augmented Generation
enabled and were shown guidelines for each case. Number of questions by specialty: 5 each
(25 total). Each response was evaluated by two clinicians.

# RAG effect on conformity to guidelines

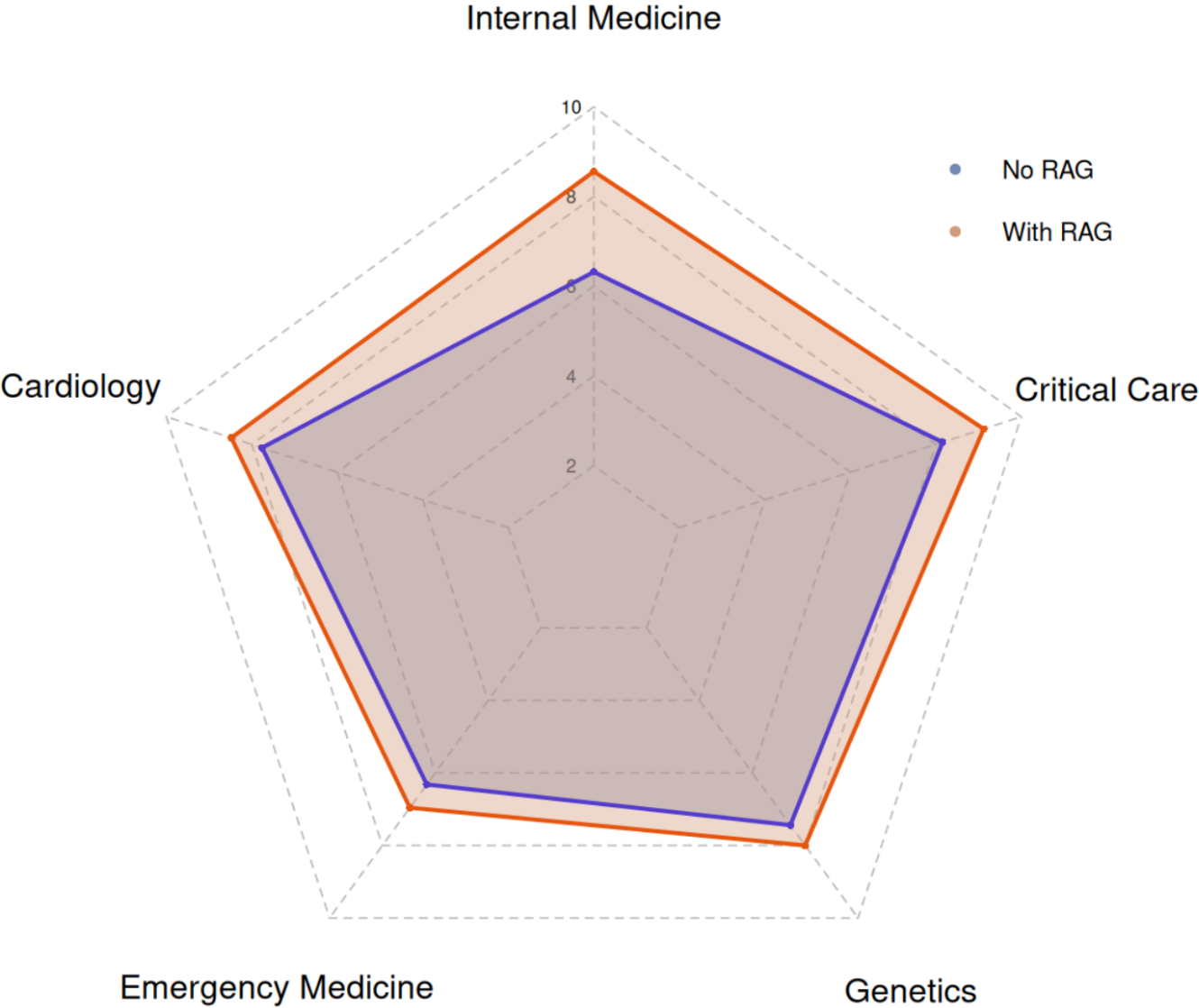

**Figure 5.** Performance change (by specialty) with and without Retrieval Augmented Generation for GPT-4 in the "Conformity to Guidelines" category.

**Tables**

| Tool name | Tool description | Notes |
|---|---|---|
| Symptom tool | Use this tool when you need to know about the patient's symptoms.<br>**The tool may be used only once**.<br>The tool does not accept any input. | Patient's history of presenting illness and physical exam. Would ordinarily be gathered at point of contact. |
| Past medical history tool | Use this tool when you need to know about the patient's past medical history.<br>**The tool may be used only once.**<br>The tool does not accept any input. | |
| Sign tool | Use this tool when you need to know about the patient's physical exam.<br>**The tool may be used only once.**<br>The tool does not accept any input. | |
| Lab investigation tool | Use this tool when you need to know about lab investigations.<br>The tool is recommended if the diagnosis is inconclusive.<br>The tool accepts a list of names of lab investigations as a string. It is very important to only order lab investigations if they are relevant.<br>**The tool must be called again with a different lab investigation if earlier results are inconclusive, not available, or insufficient.**<br>Repeating this tool with a new lab investigation is preferred before moving on to imaging studies.<br>Lab investigations themselves cannot be repeated.<br>You must specify the exact name of the lab investigation. E.g. SERUM ALBUMIN instead of just ALBUMIN.<br>Lab investigations must only be ordered from the list of available investigations provided to you.<br>Only the following lab investigations are available:<br>*<names of lab investigations>* | Names of lab investigations are pooled from all cases.<br>Multiple labs can be ordered at once.<br>Outputs the absolute value of the lab investigation alongside an interpretation in parenthesis. E.g. SERUM CREATININE: 1.1mg/dL (Elevated) |
| Imaging study tool | Use this tool when you need to know about radiological or sonographic studies.<br>The tool is recommended if the diagnosis is inconclusive.<br>The tool accepts a list of names of imaging studies as a string.<br>Only one imaging study can be ordered at a time. Start with the most relevant one.<br>**The tool must be called again with a different imaging study if earlier results are inconclusive or not available.**<br>Imaging studies must only be ordered from the list of available studies.<br>Only the following imaging studies are available:<br>*<names of imaging studies>* | Names of imaging studies are pooled from all cases.<br>Only one imaging study be ordered at a time.<br>Outputs the diagnosis statement from the imaging study. |
| ECG tool | Use this tool when you need to know about the ECG (electrocardiogram).<br>The tool is recommended regardless of how certain the diagnosis is.<br>The tool does not accept any input. | The ECG is a commonly performed, low-cost, non-invasive investigation. Therefore, the model is encouraged to utilize it. |
| Machine learning tool | Use this tool when you need to know about predictions issued by machine learning models relevant to this patient.<br>The tool is recommended to guide further testing.<br>**The tool accepts a list of names of machine learning models as a string and returns a probability value.**<br>Only one machine learning model can be used at a time.<br>Only the following machine learning models are available:<br>*<names of available outcome specific machine learning models>* | The model is only made aware of this tool in case there are relevant machine learning models available. Requires interpretation of probability. |
| Guidelines tool | Use this tool when you need to know about established guidelines.<br>Use this tool when you have a top differential diagnosis and need to know if there are any tests that can help you confirm or refute the diagnosis.<br>This tool must not be used more than once.<br>If the guidelines suggest a test you haven't ordered yet, you must order that test if it is available.<br>If the guidelines suggest a test that is not available, you must add the recommendation to your final answer.<br>You must not order tests which have already been ordered.<br>After using this tool, you must proceed to consider available treatment guidelines before giving your final answer.<br>You may not use existing knowledge to recommend a treatment unless no treatment guidelines are available.<br>**Your recommendation must be as relevant to the patient's condition as per the treatment guidelines as possible.**<br>**This tool must be used to personalize your final answer for the patient in front of you.**<br>Do not quote recommendations from guidelines verbatim.<br>The use of this tool is compulsory before issuing your final answer.<br>The tool accepts your most likely differential diagnosis as a string.<br>**Institutional guidelines take precedence over other guidelines.** | Guidelines are taken from reputable sources and distilled into two subheadings: INITIAL ASSESSMENT, and INITIAL TREATMENT. Guidelines are only shown to the model if the diagnosis provided by the model is correct. |

**Table 1. Tools**
Tool names and descriptions as they are inserted into the initial prompt for the LLM.
For tools that expect an input, italicized text is replaced by names of investigations that are pooled from all available cases at runtime. These investigations represent the degrees-of-freedom within which the model may act.
Text in bold represents safety measures to prevent the model from getting stuck in place, or specialized directions for the model that direct the model to modify its downstream output.
Lines of text in the description are separated for clarity.

| Metric | Correctness of final answer | Judicious use of tools | Conformity to guidelines |
|---|---|---|---|
| **Gemini Pro** | | | |
| **Cardiology** | 3·3 | 3 | 3·5 |
| **Critical Care** | 5·3 | 3·9 | 3·4 |
| **Emergency Medicine** | 4 | 2·6 | 2·7 |
| **Genetics** | 1·6 | 0·4 | 1 |
| **Internal Medicine** | 2·7 | 1·8 | 2·2 |
| **Claude 3 Opus** | | | |
| **Cardiology** | 8·3 | 7·5 | 8·1 |
| **Critical Care** | 10 | 8·1 | 8·1 |
| **Emergency Medicine** | 4·4 | 4·4 | 4·1 |
| **Genetics** | 9 | 6·5 | 8·4 |
| **Internal Medicine** | 7·1 | 7·5 | 7·1 |
| **GPT-4** | | | |
| **Cardiology** | 8·3 | 8·2 | 8·1 |
| **Critical Care** | 10 | 8·1 | 8·9 |
| **Emergency Medicine** | 7·5 | 7·3 | 6·2 |
| **Genetics** | 8·6 | 6·1 | 7·5 |
| **Internal Medicine** | 7·6 | 8·6 | 8·2 |

**Table 2. Model performance within framework by specialty**
Responses were manually evaluated by two expert clinicians on a scale of 1-10. Presented values are averages of all 5 cases in each specialty.

| Specialty | Score without RAG | Score with RAG | Performance uplift |
|---|---|---|---|
| Cardiology | 7·2 | 8·1 | 13% |
| Critical Care | 7·7 | 8·9 | 16% |
| Emergency Medicine | 5·4 | 6·2 | 15% |
| Genetics | 6·8 | 7·5 | 10% |
| Internal Medicine | 5·4 | 7·2 | 52% |
| Average | | | 20% |

**Table 3. Performance change for best performing model (GPT-4) with RAG**
Performance improvement in the "Conformity to guidelines" metric with and without Retrieval Augmented Generation enabled. Values listed are out of 10.

**Funding**

This work was supported in part through the computational resources and staff expertise provided by Scientific Computing at the Icahn School of Medicine at Mount Sinai and supported by the Clinical and Translational Science Awards (CTSA) grant UL1TR004419 from the National Center for Advancing Translational Sciences.

**Author contributions**

The study was conceived and designed by AV; Code was written by AV; Underlying data were analyzed and visualized by AV; Cases were collected and evaluated by J.Lampert, J.Lee, A.Sawant, DA, A.Sakhuja; Cases were evaluated by SB, EA, HG, MH, GNN; The first draft of the manuscript was written by AV; GNN supervised the project and acquired funding. AV and GNN had access to and verified the data. All authors had access to the data presented in the study, contributed to critical revisions, and approved the final version of the manuscript as well as the decision for submission.

**Competing Interests**

Dr. Nadkarni reports consultancy agreements with AstraZeneca, BioVie, GLG Consulting, Pensieve Health, Reata, Renalytix, Siemens Healthineers, and Variant Bio; research funding from Goldfinch Bio and Renalytix; honoraria from AstraZeneca, BioVie, Lexicon, Daiichi Sankyo, Meanrini Health and Reata; patents or royalties with Renalytix; owns equity and stock options in Pensieve Health and Renalytix as a scientific cofounder; owns equity in Verici Dx; has received financial compensation as a scientific board member and advisor to Renalytix; serves on the advisory board of Neurona Health; and serves in an advisory or leadership role for Pensieve Health and Renalytix. Dr. Khera is an Associate Editor of JAMA. He receives support from the National Heart, Lung, and Blood Institute of the National Institutes of Health (under award K23HL153775) and the Doris Duke Charitable Foundation (under award, 2022060). He also receives research support, through Yale, from Bristol-Myers Squibb, Novo Nordisk, and BridgeBio. He is a coinventor of U.S. Provisional Patent Applications 63/177,117, 63/428,569, 63/346,610, 63/484,426, 63/508,315, and 63/606,203. He is a co-founder of Ensight-AI and Evidence2Health, health platforms to improve cardiovascular diagnosis and evidence-based cardiovascular care. Dr. Lampert reports a consultancy agreement with Viz.ai. All other authors have reported that they have no relationships relevant to the contents of this paper to disclose.